\documentclass{article} 
\usepackage{iclr2027_conference,times}

\usepackage[T1]{fontenc}

\usepackage{amsmath,amsfonts,bm}

\def\1{\bm{1}}

\def\vtheta{{\bm{\theta}}}

\def\vc{{\bm{c}}}

\def\vg{{\bm{g}}}

\def\vm{{\bm{m}}}

\def\vr{{\bm{r}}}

\def\vv{{\bm{v}}}

\def\mA{{\bm{A}}}
\def\mB{{\bm{B}}}

\def\mG{{\bm{G}}}
\def\mH{{\bm{H}}}
\def\mI{{\bm{I}}}

\def\mP{{\bm{P}}}
\def\mQ{{\bm{Q}}}

\def\mU{{\bm{U}}}

\def\mW{{\bm{W}}}
\def\mX{{\bm{X}}}
\def\mY{{\bm{Y}}}

\def\mSigma{{\bm{\Sigma}}}

\DeclareMathAlphabet{\mathsfit}{\encodingdefault}{\sfdefault}{m}{sl}
\SetMathAlphabet{\mathsfit}{bold}{\encodingdefault}{\sfdefault}{bx}{n}

\usepackage{graphicx}
\usepackage{xcolor}
\usepackage{subcaption}
\usepackage{booktabs}
\usepackage{multirow}
\usepackage[ruled,vlined]{algorithm2e}
\usepackage{hyperref}
\usepackage{url}

\DeclareMathOperator{\diag}{diag}
\DeclareMathOperator{\orth}{\text{\normalfont\itshape Orth}}
\newcommand{\rownorm}{\mathcal{N}_{r}}
\newcommand{\colnorm}{\mathcal{N}_{c}}
\newcommand{\vecnorm}{\mathcal{N}}

\title{MeqMuon: Matrix-Equilibrating Muon for LLM Pretraining}

\author{    
Chang-Wei Shi,
Xu Wang,
Wu-Jun Li \thanks{Corresponding author.} \\
National Key Laboratory for Novel Software Technology, \\
School of Computer Science, Nanjing University, P. R. China\\
\texttt{
\{shicw, wangxu0328\}@smail.nju.edu.cn, liwujun@nju.edu.cn} 
}

\iclrfinalcopy 
\begin{document}

\maketitle

\begin{abstract}
The success of large language models (LLMs) has been accompanied by continued
growth in model size and pretraining costs.
Muon offers high accuracy and training efficiency in LLM pretraining.
Recent work introduces row-wise normalization into Muon to balance update
magnitudes and improve pretraining performance.
However, row-wise normalization alone cannot accommodate different imbalance
patterns in update matrices.
In this paper, we propose an improved Muon optimizer, called \underline{m}atrix-\underline{eq}uilibrating Muon~(MeqMuon), for LLM pretraining.
MeqMuon balances both row and column magnitudes through
normalization that can be automatically tailored to different imbalance patterns without manual intervention.
Moreover, MeqMuon eliminates the need to store AdamW's second-moment estimates,
reducing optimizer-state memory usage.
Empirical results demonstrate that MeqMuon achieves better convergence
performance than AdamW, Muon, and other baselines in LLM pretraining.
\end{abstract}

\section{Introduction}
\label{sec:introduction}

The rapid development of large language models (LLMs) has enabled advances
across diverse domains, such as code generation \citep{chen2021evaluating,deepseek2026v4},
mathematical problem solving \citep{shao2025deepseekmathv2}, and protein
structure prediction \citep{lin2023evolutionary,candido2026language}.
These advances have been accompanied by continued growth in model size.
For example, DeepSeek-V4-Pro \citep{deepseek2026v4} has 1.6 trillion total
parameters and was pretrained on 33 trillion tokens, while Kimi K3
\citep{kimiteam2026k3} reaches 2.8 trillion total parameters.
Pretraining models at such scales incurs substantial cost.
Efficient optimizer design is therefore important for improving model
performance and reducing training cost
\citep{loshchilov2019adamw,shazeer2018adafactor,dettmers2022eightbit,
chen2023lion,jordan2024muon,liu2025scalable,
pagliardini2025ademamix,zhang2025adammini,zhao2024galore,zhu2025apollo}.

Adam \citep{kingma2015adam} has long been the dominant
optimizer for LLM pretraining.
It uses first- and second-moment estimates of gradients to compute
coordinate-wise adaptive updates.
AdamW \citep{loshchilov2019adamw}, a variant of Adam with decoupled weight decay,
has been used to pretrain LLMs such as Llama 2 \citep{touvron2023llama2} and
DeepSeek-V3 \citep{deepseek2024v3}.
Recently, Muon \citep{jordan2024muon,liu2025scalable} has demonstrated substantial gains over AdamW
in training efficiency and model accuracy.
Its effectiveness has also been demonstrated in pretraining frontier LLMs such as
DeepSeek-V4 \citep{deepseek2026v4}, Kimi K2 \citep{kimiteam2025k2},
and Kimi K3 \citep{kimiteam2026k3}.
Muon applies different update rules to 2D weights in hidden layers
and the remaining parameters.
For 2D weights in hidden layers, it orthogonalizes momentum matrices
to bring their nonzero singular values toward one.
The remaining parameters, such as token embedding matrices, language modeling
(LM) heads, and 1D parameters, are updated with AdamW.

Normalization is a widely used technique in optimizer design.
The convergence properties of normalized gradient methods have been studied
theoretically
\citep{levy2016power,murray2019revisiting,zhang2020clipping,
cutkosky2020momentum,zhao2024stochastic,yang2024batch,sun2025revisiting}.
In LLM pretraining, normalization has been incorporated into optimizers in
different forms.
AdamW \citep{loshchilov2019adamw} normalizes first-moment estimates coordinate-wise
using the square roots of second-moment estimates.
Muon \citep{jordan2024muon,liu2025scalable} normalizes singular values by
approximately orthogonalizing momentum matrices for 2D weights in hidden layers.
\mbox{SCALE~\citep{glentis2026scale}} normalizes the update vector associated
with each output dimension for all 2D weights.
\citet{li2025normuon} observe substantial imbalance among row magnitudes
in some of Muon's orthogonalized updates.
They propose NorMuon, which applies Adam-style row-wise normalization to mitigate
this imbalance and achieves better convergence performance than Muon in LLM
pretraining.
However, our observation shows that Muon's orthogonalized updates can exhibit
imbalance in either row or column magnitudes.
For example, some updates may have relatively balanced row magnitudes but
imbalanced column magnitudes, while others may have relatively balanced column
magnitudes but imbalanced row magnitudes.
Row-wise normalization alone therefore cannot accommodate different imbalance
patterns in update matrices.

In this paper, we propose an improved Muon optimizer, called \underline{m}atrix-\underline{eq}uilibrating Muon~(\mbox{MeqMuon}), for LLM pretraining.
The main contributions of this paper are outlined as follows:
\begin{itemize}
    \item[$\bullet$] We identify different imbalance patterns of row and column magnitudes in Muon.
    Muon's orthogonalized updates can exhibit imbalance predominantly in either
    row or column magnitudes. Unorthogonalized momentum matrices for
    the remaining 2D parameters exhibit imbalance in both row and column magnitudes.
    \item[$\bullet$] We propose MeqMuon, which balances both row and column magnitudes
    through normalization that can be automatically tailored to different imbalance patterns without manual intervention. MeqMuon selects row-wise or column-wise normalization for
    Muon's orthogonalized updates and applies two-sided normalization to
    unorthogonalized momentum matrices for the remaining 2D parameters.
    \item[$\bullet$] MeqMuon eliminates second-moment storage and reduces
    optimizer-state memory usage by replacing AdamW updates with normalized momentum for the remaining
    parameters.
    \item[$\bullet$] Empirical results demonstrate that MeqMuon achieves better
    convergence performance than AdamW, Muon, and other baselines in LLM pretraining.
\end{itemize}

\section{Preliminaries}
\label{sec:preliminaries}

\paragraph{Problem formulation.}
LLM pretraining is formulated as the following
optimization problem
\begin{equation}
\min_{\vtheta}\;\mathcal{L}(\vtheta),
\label{eq:objective}
\end{equation}
where $\mathcal{L}$ is the training loss and $\vtheta$ denotes the collection
of all model parameter tensors.
LLM parameters are predominantly 1D or 2D.
Higher-dimensional parameter tensors are reshaped or partitioned into 2D matrices
for optimization in Muon implementations \citep{jordan2024muon}.
We therefore focus on 1D and 2D parameters in this paper.

\paragraph{AdamW.}
Adam \citep{kingma2015adam} maintains first- and
second-moment estimates of the minibatch gradient
$\vg_t=\nabla_{\vtheta}\mathcal{L}(\vtheta_t)$ using
exponential moving averages, where $t$ denotes the iteration number.
Starting from $\vm_0=\vv_0=\mathbf 0$, Adam follows the update rules below:
\begin{equation}
\begin{gathered}
\begin{aligned}
\vm_t&=\beta_1\vm_{t-1}+(1-\beta_1)\vg_t,
&\widehat{\vm}_t&=\frac{\vm_t}{1-\beta_1^t},\\
\vv_t&=\beta_2\vv_{t-1}+(1-\beta_2)(\vg_t\odot\vg_t),
&\widehat{\vv}_t&=\frac{\vv_t}{1-\beta_2^t},
\end{aligned}\\[2pt]
\vtheta_{t+1}=\vtheta_t
-\eta_t\frac{\widehat{\vm}_t}{\sqrt{\widehat{\vv}_t}+\epsilon}.
\end{gathered}
\label{eq:adam-moments}
\end{equation}
Here $\beta_1,\beta_2\in[0,1)$ are exponential decay rates, $\eta_t$ is the
learning rate, and $\epsilon>0$ is a constant for numerical stability.
The symbol $\odot$ denotes element-wise multiplication.

In LLM pretraining, AdamW \citep{loshchilov2019adamw} is widely used in place
of Adam. AdamW modifies Adam through decoupled weight decay:
\begin{equation}
\vtheta_{t+1}=(1-\eta_t\lambda)\vtheta_t
-\eta_t\frac{\widehat{\vm}_t}{\sqrt{\widehat{\vv}_t}+\epsilon},
\label{eq:adamw-update}
\end{equation}
where $\lambda\geq 0$ is the weight decay coefficient.

\paragraph{Muon.}
Muon \citep{jordan2024muon} orthogonalizes momentum matrices
for 2D weights in hidden layers.
A weight matrix $\mW_t\in\mathbb{R}^{m\times n}$ in a hidden layer
with minibatch gradient $\mG_t$ is updated as
\begin{equation}
\mB_t=\mu\mB_{t-1}+\mG_t,\qquad
\mU_t=\orth(\mB_t),\qquad
\mW_{t+1}=\mW_t-\eta_t\mU_t,
\label{eq:muon-update}
\end{equation}
where $\mB_{-1}=\mathbf 0$, $\mB_t$ is the momentum matrix, and
$\mu\in[0,1)$ is the momentum coefficient.
For $\mH\in\mathbb{R}^{m\times n}$ with SVD
$\mH=\mP\mSigma\mQ^\top$ restricted to nonzero singular values,
$\orth(\mH)$ is defined as $\mP\mQ^\top$.
In practice, $\orth$ is approximated by $K$ Newton--Schulz (NS) iterations.
For a nonzero input $\mH$, the quintic iteration starts from
$\mX_0=\mH/\|\mH\|_F$ and takes the form
\begin{equation}
\mA_k=\mX_k\mX_k^\top,\qquad
\mX_{k+1}=a\mX_k+(b\mA_k+c\mA_k^2)\mX_k,\qquad k=0,\ldots,K-1.
\label{eq:ns}
\end{equation}
Existing works \citep{jordan2024muon,liu2025scalable} commonly use
$(a,b,c)=(3.4445,-4.7750,2.0315)$ and $K=5$.
Muon applies these orthogonalized updates only to 2D weights in hidden layers,
while the remaining parameters, such as token embedding matrices, LM heads, and
1D parameters, are updated with AdamW.

For large-scale LLM pretraining, \citet{liu2025scalable} introduced
decoupled weight decay and root mean square (RMS) alignment into Muon.
These modifications were later adopted in Kimi K2
\citep{kimiteam2025k2} and DeepSeek-V4 \citep{deepseek2026v4}.
With these modifications, the update rule in \eqref{eq:muon-update}
becomes
\begin{equation}
\mB_t=\mu\mB_{t-1}+\mG_t,\quad
\mU_t=\orth(\mB_t),\quad
\mW_{t+1}=(1-\eta_t\lambda)\mW_t
-\eta_t\rho\sqrt{\max(m,n)}\,\mU_t,
\label{eq:moonlight-update}
\end{equation}
where $\lambda\geq 0$ is the weight decay coefficient.
$\sqrt{\max(m,n)}$ compensates for the effect of matrix shape on the update RMS.
The RMS alignment coefficient $\rho>0$ sets the target update RMS and is chosen
to approximately match that of AdamW.
Moonlight and Kimi K2 use $\rho=0.2$
\citep{liu2025scalable,kimiteam2025k2}, while DeepSeek-V4 uses $\rho=0.18$
\citep{deepseek2026v4}.

\section{Method}
\label{sec:method}

In this section, we first characterize the different imbalance patterns in row
and column magnitudes and then introduce our proposed method, MeqMuon.

\subsection{Imbalance patterns}
\label{sec:diagnostics}
To characterize imbalance in matrix updates, we first define statistics for their row and column magnitudes. We follow the matrix-layout
conventions used in PyTorch implementations.
For linear weight matrices, rows correspond to output features and columns
correspond to input features.
For token embedding matrices and LM heads, rows correspond to vocabulary entries
and columns correspond to hidden features.
For a matrix $\mX\in\mathbb{R}^{m\times n}$, the row-wise and column-wise
root mean square (RMS) magnitudes are collected in vectors
$\vr\in\mathbb{R}^{m}$ and $\vc\in\mathbb{R}^{n}$, respectively,
with entries defined as
\begin{equation}
r_i=\frac{\|\mX_{i,:}\|_2}{\sqrt n},\quad i=1,\ldots,m,
\qquad
c_j=\frac{\|\mX_{:,j}\|_2}{\sqrt m},\quad j=1,\ldots,n.
\label{eq:row-col-rms}
\end{equation}
Imbalance in row and column magnitudes is quantified by the coefficients of
variation (CVs) of $\vr$ and $\vc$, denoted by $\gamma_r$ and $\gamma_c$,
respectively.
Each CV is defined as the ratio of the standard deviation to the mean:
\begin{equation}
\begin{aligned}
\bar r&=\frac{1}{m_+}\sum_{i:\,r_i>0}r_i,&
\gamma_r&=\frac{\sqrt{\frac{1}{m_+}\sum_{i:\,r_i>0}(r_i-\bar r)^2}}{\bar r},\\
\bar c&=\frac{1}{n_+}\sum_{j:\,c_j>0}c_j,&
\gamma_c&=\frac{\sqrt{\frac{1}{n_+}\sum_{j:\,c_j>0}(c_j-\bar c)^2}}{\bar c},
\end{aligned}
\label{eq:cv}
\end{equation}
where $m_+$ and $n_+$ denote the numbers of nonzero entries in $\vr$ and $\vc$,
respectively. Zero rows and columns are excluded from the corresponding CV
calculations. In our implementation, RMS values at or below $\delta=10^{-7}$
are treated as zero. The CVs are dimensionless and invariant to uniform scaling.
Larger CVs indicate greater imbalance.
Accordingly, matrix equilibration seeks to balance row and column magnitudes,
as reflected by lower row and column CVs.

Using these statistics, we first examine Muon's orthogonalized updates for 2D weights in hidden layers. These update matrices can exhibit
imbalance predominantly in either row or column magnitudes.
For example, some updates may have relatively balanced row magnitudes but
imbalanced column magnitudes, while others may have relatively balanced column
magnitudes but imbalanced row magnitudes.
As a concrete example, we examine Transformer layer 4 of Llama-350M at training
step 1,144. For each 2D weight in this layer, we record the saved post-NS update
produced after 5 NS iterations, corresponding to $\mU_t$ in
\eqref{eq:muon-update}. Table~\ref{tab:row-column-imbalance} shows the row
and column CVs of these updates. The orthogonalized update matrices associated 
with \texttt{q\_proj}, \texttt{k\_proj}, \texttt{v\_proj}, \texttt{gate\_proj}, 
and \texttt{up\_proj} have relatively balanced column magnitudes but imbalanced 
row magnitudes, whereas those associated with \texttt{o\_proj} and \texttt{down\_proj} 
exhibit the opposite pattern. Thus, after 5 NS iterations, either the row magnitudes or the column magnitudes can be nearly balanced, while the other remains imbalanced, even within the same
Transformer layer.
Additional imbalance patterns across different models and
training steps are reported in Appendix~\ref{app:imbalance}.

\begin{table}[!t]
  \caption{Row and column CVs of Muon's orthogonalized updates after 5 NS iterations.}
  \label{tab:row-column-imbalance}
  \centering
  \normalsize
  \setlength{\tabcolsep}{6pt}
  \renewcommand{\arraystretch}{1.12}
  \begin{tabular}{lccc}
    \toprule
    Parameter matrix & Shape & Row CV & Column CV \\
    \midrule
    \texttt{q\_proj}    & $1024\times1024$ & \textbf{0.194} & 0.027 \\
    \texttt{k\_proj}    & $1024\times1024$ & \textbf{0.180} & 0.024 \\
    \texttt{v\_proj}    & $1024\times1024$ & \textbf{0.195} & 0.041 \\
    \texttt{o\_proj}    & $1024\times1024$ & 0.014 & \textbf{0.091} \\
    \texttt{gate\_proj} & $2736\times1024$ & \textbf{0.161} & 0.017 \\
    \texttt{up\_proj}   & $2736\times1024$ & \textbf{0.159} & 0.016 \\
    \texttt{down\_proj} & $1024\times2736$ & 0.015 & \textbf{0.181} \\
    \bottomrule
  \end{tabular}
\end{table}

\begin{figure}[!t]
  \centering
  \includegraphics[width=\linewidth]{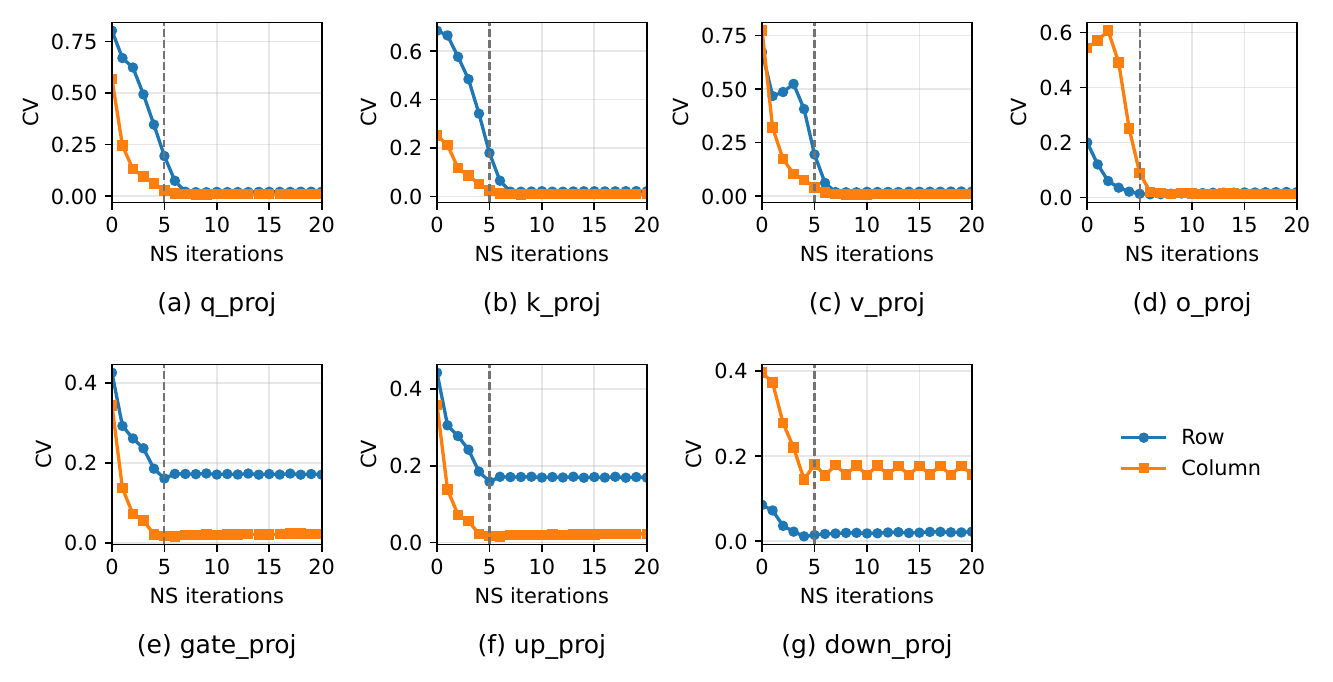}
  \caption{Row and column CV trajectories during NS iterations.}
  \label{fig:ns-trajectories}
\end{figure}

To examine how imbalance in row and column magnitudes changes during orthogonalization,
we replay the NS iteration in \eqref{eq:ns} using the saved pre-NS matrices and
track both CVs from $k=0$ to $k=20$. Figure~\ref{fig:ns-trajectories} shows the
resulting trajectories. For the square matrices, the row and column CVs decrease
at different rates during the first few iterations. For \texttt{v\_proj}, for
example, the column CV initially exceeds the row CV, but this ordering reverses
after 5 iterations: $(\gamma_r,\gamma_c)$ changes from $(0.673,0.773)$ at $k=0$
to $(0.195,0.041)$ at $k=5$. By $k=20$, both row and column CVs are close to zero
for the square matrices. For the rectangular matrices, one CV approaches zero
as the NS iterations proceed, while the other remains large.
These observations can be understood from the constraints imposed by
orthogonalization. Let $\mX\in\mathbb{R}^{m\times n}$ be full-rank, and let
$\mY=\orth(\mX)$ denote its exact orthogonalization. When $m>n$,
$\mY^\top\mY=\mI_n$, so all
column magnitudes are equal and $\gamma_c=0$, while the row magnitudes remain
unconstrained. When $m<n$, $\mY\mY^\top=\mI_m$, so all row magnitudes are equal
and $\gamma_r=0$, while the column magnitudes remain unconstrained. Thus, for
rectangular matrices, exact orthogonalization constrains only one side, allowing
one-sided imbalance to persist. For square matrices, exact orthogonalization gives $\gamma_r=\gamma_c=0$. The residual imbalance
observed after 5 NS iterations therefore reflects finite-iteration approximation
error, with the row and column CVs decreasing at different rates.

For the remaining 2D parameters, we separately accumulate unorthogonalized
momentum matrices for \texttt{embed\_tokens} and \texttt{lm\_head} in Llama-350M.
Table~\ref{tab:remaining-momentum-imbalance} reports their row and column CVs at
training step 1,144. Each matrix exhibits imbalance in both row and column
magnitudes. Although their row CVs are substantially larger, their column CVs
also remain non-negligible. Since these momentum matrices are not orthogonalized,
no orthogonality condition enforces equal row or column magnitudes.

\begin{table}[!t]
  \caption{Row and column CVs of unorthogonalized momentum matrices for the
  remaining 2D parameters.}
  \label{tab:remaining-momentum-imbalance}
  \centering
  \normalsize
  \setlength{\tabcolsep}{6pt}
  \renewcommand{\arraystretch}{1.12}
  \begin{tabular}{lccc}
    \toprule
    Parameter matrix & Shape & Row CV & Column CV \\
    \midrule
    \texttt{embed\_tokens} & $32000\times1024$ & \textbf{1.225} & 0.185 \\
    \texttt{lm\_head}      & $32000\times1024$ & \textbf{1.384} & 0.493 \\
    \bottomrule
  \end{tabular}
\end{table}

NorMuon \citep{li2025normuon} applies row-wise normalization to address the
row imbalance observed in Muon's orthogonalized updates.
Our observation shows that Muon's orthogonalized updates can exhibit imbalance
predominantly in either row or column magnitudes, whereas unorthogonalized
momentum matrices for the remaining 2D parameters exhibit imbalance in both row
and column magnitudes. Row-wise normalization alone therefore cannot accommodate
these different imbalance patterns. To address this limitation, we propose
MeqMuon, which balances both row and column magnitudes through normalization tailored
to these patterns.

\subsection{MeqMuon}
\label{sec:hidden}
\label{sec:operators}
\label{sec:equilibration}
\label{sec:auxiliary}
\label{sec:scale-cost}

MeqMuon maintains one momentum buffer for each parameter and applies
normalization to the updates according to parameter type. It uses adaptive
one-sided normalization for 2D weights in hidden layers, two-sided normalization
for the remaining 2D parameters, and global-RMS normalization for each 1D parameter.

For a matrix $\mX\in\mathbb{R}^{m\times n}$, we define RMS-based row-wise
and column-wise normalization using the RMS vectors $\vr$ and $\vc$ in
\eqref{eq:row-col-rms}:
\begin{equation}
\rownorm(\mX)=\diag(\vr^{-1})\mX,
\qquad
\colnorm(\mX)=\mX\diag(\vc^{-1}),
\label{eq:operators}
\end{equation}
where $\diag(\vr^{-1})$ denotes a diagonal matrix with the diagonal elements being $\vr^{-1}$. These operators rescale each nonzero row or column to unit RMS, respectively.

For each parameter, we initialize the momentum as $\mB_{-1}=\mathbf{0}$
and update it as follows:
\begin{equation}
\mB_t=\mu\mB_{t-1}+\mG_t,
\label{eq:momentum}
\end{equation}
where $\mG_t$ is the corresponding gradient and $\mu$ is the momentum
coefficient.

For 2D weights in hidden layers, we follow Muon to compute the orthogonalized
update $\mU_t=\orth(\mB_t)$.
We then select row-wise or column-wise normalization by comparing the row
and column CVs of $\mU_t$:
\begin{equation}
\widetilde{\mU}_t=
\begin{cases}
\rownorm(\mU_t),&if\hspace{0.2cm} \gamma_r(\mU_t)\geq\gamma_c(\mU_t),\\
\colnorm(\mU_t),&if\hspace{0.2cm} \gamma_r(\mU_t)<\gamma_c(\mU_t).
\end{cases}
\label{eq:hidden}
\end{equation}
The normalization direction is selected independently for each matrix at every
optimization step.

For the remaining 2D parameters, such as token embedding matrices and LM heads,
we replace Muon's AdamW updates with two-sided normalization of the
unorthogonalized momentum:
\begin{equation}
\widetilde{\mU}_t=\colnorm\!\left(\rownorm(\mB_t)\right).
\label{eq:aux}
\end{equation}
The two-sided normalization consists of both row-wise and column-wise
normalization. 

\begin{table}[!t]
  \caption{Row and column CVs before and after normalization.}
  \label{tab:normalization-cv}
  \centering
  \normalsize
  \setlength{\tabcolsep}{6pt}
  \renewcommand{\arraystretch}{1.12}
  \begin{tabular}{lcccc}
    \toprule
    & \multicolumn{2}{c}{Before} & \multicolumn{2}{c}{After} \\ 
    \cmidrule(lr){2-3} \cmidrule(lr){4-5}
    Parameter matrix & Row CV & Column CV & Row CV & Column CV \\ 
    \midrule
    \texttt{q\_proj} & 0.194 & 0.027 & $<0.001$ & 0.027 \\
    \texttt{k\_proj} & 0.180 & 0.024 & $<0.001$ & 0.023 \\
    \texttt{v\_proj} & 0.195 & 0.041 & $<0.001$ & 0.041 \\
    \texttt{o\_proj} & 0.014 & 0.091 & 0.014 & $<0.001$ \\
    \texttt{gate\_proj} & 0.161 & 0.017 & $<0.001$ & 0.018 \\
    \texttt{up\_proj} & 0.159 & 0.016 & $<0.001$ & 0.018 \\
    \texttt{down\_proj} & 0.015 & 0.181 & 0.015 & $<0.001$ \\
    \midrule
    \texttt{embed\_tokens} & 1.225 & 0.185 & 0.008 & $<0.001$ \\
    \texttt{lm\_head} & 1.384 & 0.493 & 0.114 & $<0.001$ \\
    \bottomrule
  \end{tabular}
\end{table}

Table~\ref{tab:normalization-cv} compares the row and column CVs of the matrices
in Table~\ref{tab:row-column-imbalance}
and Table~\ref{tab:remaining-momentum-imbalance} before and after the corresponding
normalization. For Muon's orthogonalized updates, row-wise or column-wise
normalization reduces the larger CV to nearly zero while keeping the other
CV small. For the remaining 2D parameters, two-sided normalization substantially
reduces both row and column CVs.
MeqMuon thus balances both row and column magnitudes through normalization tailored to different imbalance patterns.

For 1D parameters, such as RMSNorm weights
\citep{zhang2019rmsnorm} and bias vectors, we normalize the momentum $\mB_t\in\mathbb{R}^d$ by its global RMS:
\begin{equation}
\vecnorm(\mB_t)=\frac{\mB_t}{\|\mB_t\|_2/\sqrt d}.
\label{eq:vector}
\end{equation}

The parameter update rule in MeqMuon is given by
\begin{equation}
\mW_{t+1}=(1-\eta_t\lambda)\mW_t
-\rho\eta_t\widetilde{\mU}_t,
\label{eq:meqmuon-update}
\end{equation}
where the first term applies decoupled weight decay and the second term scales
the normalized update by $\rho\eta_t$.
Either row-wise or column-wise normalization makes the update RMS independent
of matrix shape. Thus, $\sqrt{\max(m,n)}$ in \eqref{eq:moonlight-update}
is no longer needed. Algorithm~\ref{alg:meqmuon} summarizes the update rules of MeqMuon.
\begin{algorithm}[t]
\caption{MeqMuon}
\label{alg:meqmuon}
\DontPrintSemicolon
\KwIn{Initial parameters $\{\mW_0\}$, total training steps $T$,
learning rates $\{\eta_t\}_{t=0}^{T-1}$, coefficients $\mu,\rho,\lambda$;}
\KwOut{Parameters $\{\mW_T\}$;}
Initialize $\mB_{-1}\leftarrow\mathbf{0}$ for every parameter;\;
\For{$t=0,\ldots,T-1$}{
  \ForEach{trainable parameter $\mW_t$}{
    Obtain its gradient $\mG_t$;\;
    $\mB_t\leftarrow\mu\mB_{t-1}+\mG_t$;\;
    \uIf{$\mW_t$ is a 2D weight in a hidden layer}{
      $\mU_t\leftarrow\orth(\mB_t)$;\;
      \eIf{$\gamma_r(\mU_t)\geq\gamma_c(\mU_t)$}{
        $\widetilde{\mU}_t\leftarrow\rownorm(\mU_t)$;\;
      }{
        $\widetilde{\mU}_t\leftarrow\colnorm(\mU_t)$;\;
      }
    }\uElseIf{$\mW_t$ is one of the remaining 2D parameters}{
      $\widetilde{\mU}_t\leftarrow\colnorm(\rownorm(\mB_t))$;\;
    }\Else{
      $\widetilde{\mU}_t\leftarrow\vecnorm(\mB_t)$;\;
    }
    $\mW_{t+1}\leftarrow(1-\eta_t\lambda)\mW_t
      -\rho\eta_t\widetilde{\mU}_t$;\;
  }
}
\end{algorithm}

\begin{table}[t]
  \caption{Validation PPL for pretraining different models ($\downarrow$).}
  \label{tab:experiment-main}
  \centering
  \normalsize
  \setlength{\tabcolsep}{8.5pt}
  \renewcommand{\arraystretch}{1.12}
  \begin{tabular}{lcccccc}
    \toprule
    & \multicolumn{3}{c}{Llama} & \multicolumn{2}{c}{SmolLM2} & Qwen2 \\
    \cmidrule(lr){2-4}\cmidrule(lr){5-6}\cmidrule(lr){7-7}
    Model scale & 60M & 130M & 350M & 135M & 360M & 0.5B \\
    Token budget & 1.2B & 2.7B & 7.4B & 2.7B & 7.2B & 9.9B \\
    \midrule
    AdamW & 37.40 & 24.46 & 17.02 & 24.88 & 18.46 & 19.65 \\
    SCALE & 58.38 & 33.53 & 19.87 & -- & -- & -- \\
    Muon & 29.88 & 21.64 & 16.04 & 22.81 & 17.28 & 18.79 \\
    NorMuon & 29.72 & 21.57 & 15.98 & 22.83 & 17.20 & 18.76 \\
    MeqMuon & \textbf{29.53} & \textbf{21.38} & \textbf{15.94} & \textbf{22.80} & \textbf{17.16} & \textbf{18.63} \\
    \bottomrule
  \end{tabular}
\end{table}

\section{Experiments}
\label{sec:experiments}
In this section, we evaluate the performance of MeqMuon and other baselines
for LLM pretraining.
All the experiments are conducted on NVIDIA RTX A6000 GPUs.
All the methods are implemented on PyTorch 2.6.0 with CUDA 12.4 and
Transformers 4.57.6, using the DistributedDataParallel (DDP) framework.
We use \texttt{torch.compile},
BF16 automatic mixed-precision training with FP32 master weights,
and scaled dot-product attention with the Flash backend enabled.

We evaluate Llama \citep{touvron2023llama2}, SmolLM2
\citep{benallal2025smollm2}, and Qwen2 \citep{yang2024qwen2} models.
Llama uses the T5-base tokenizer \citep{raffel2020t5}, while the other models use their native
tokenizers.
Llama uses separate weights for the token embeddings
and the LM head, whereas SmolLM2 and Qwen2 share these weights.
Details of the model architectures are provided in
Appendix~\ref{app:implementation-settings}.
All models are trained from random initialization.
We use English C4 \citep{raffel2020t5} for pretraining,
with text packed into fixed-length sequences.
Following the Chinchilla compute-optimal training rule
\citep{hoffmann2022training}, we set the training token budget to
20 times the number of model parameters.
We report perplexity (PPL) on the validation set, defined as the
exponential of the average per-token cross-entropy loss.

We compare MeqMuon with AdamW \citep{loshchilov2019adamw},
SCALE \citep{glentis2026scale}, Muon \citep{liu2025scalable},
and NorMuon \citep{li2025normuon}.
We use PyTorch's built-in AdamW.
SCALE and NorMuon are based on their official implementations.
Muon follows
Moonlight's implementation \citep{liu2025scalable} as shown in
\eqref{eq:moonlight-update}.
We set the sequence length to 1,024 and the global batch size to 512.
All experiments use decoupled weight decay
with a coefficient of 0.1. 
Gradient clipping and activation checkpointing are disabled.
We use linear learning rate warmup for the first 5\% of updates,
followed by cosine learning rate decay.
AdamW and the AdamW updates for the remaining parameters in Muon and NorMuon
use $(\beta_1,\beta_2)=(0.9,0.95)$.
For 2D weights in hidden layers, Muon, NorMuon, and MeqMuon use
Nesterov momentum with coefficient $\mu=0.95$.
They use 5 NS iterations with coefficients
$(a,b,c)=(3.4445,-4.7750,2.0315)$ and RMS alignment coefficient $\rho=0.2$. SCALE is evaluated only on Llama models, because it updates the token
embedding and the LM head in different ways, which is incompatible with
the shared embedding and LM head weights in SmolLM2 and Qwen2.

\paragraph{Convergence performance.}

\begin{figure}[t]
  \centering
  \includegraphics[width=\linewidth]{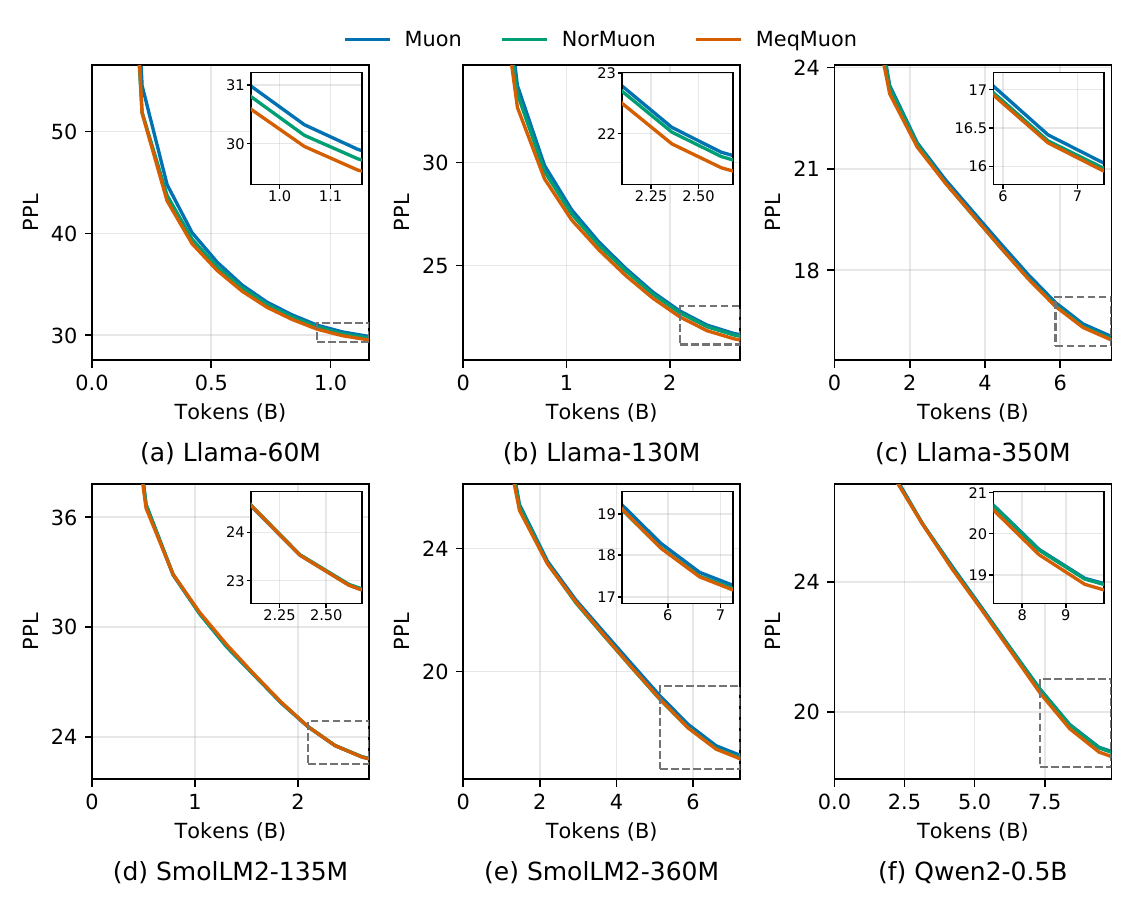}
  \caption{Validation PPL curves during pretraining.}
  \label{fig:experiment-convergence}
\end{figure}

Table~\ref{tab:experiment-main} reports the final validation PPL of different
optimizers across the evaluated models, and
Figure~\ref{fig:experiment-convergence} compares the validation PPL curves of
Muon, NorMuon, and MeqMuon during pretraining. Muon, NorMuon, and MeqMuon
achieve substantially lower final PPL than AdamW and SCALE, while MeqMuon
obtains the best final PPL for every reported model scale. Its validation PPL
curves also generally remain below those of Muon and NorMuon during training,
demonstrating consistently improved convergence.

\paragraph{Memory overhead.}
Let $N_h$ denote the total number of elements across all 2D weights in hidden
layers, and let $N_a$ denote the number of elements in the remaining parameters.
Let $R_h$ denote the total number of rows across all 2D weights in hidden
layers. Muon stores $N_h$ momentum values and $2N_a$ AdamW first- and second-moment values. Compared with the memory overhead of Muon, NorMuon additionally stores $R_h$ row-wise second-moment values. MeqMuon
stores only $N_h+N_a$ momentum values. Table~\ref{tab:experiment-efficiency} shows the optimizer-state memory of different optimizers, which confirms the predicted savings of MeqMuon over Muon and NorMuon.
For example, on Qwen2-0.5B, MeqMuon reduces the optimizer-state memory by 21.6\% compared with Muon and NorMuon.

\begin{table}[!t]
  \caption{Optimizer-state memory of different optimizers (MiB, $\downarrow$).}
  \label{tab:experiment-efficiency}
  \centering
  \normalsize
  \setlength{\tabcolsep}{8.5pt}
  \renewcommand{\arraystretch}{1.12}
  \begin{tabular}{lcccccc}
    \toprule
    & \multicolumn{3}{c}{Llama} & \multicolumn{2}{c}{SmolLM2} & Qwen2 \\
    \cmidrule(lr){2-4}\cmidrule(lr){5-6}\cmidrule(lr){7-7}
    & 60M & 130M & 350M & 135M & 360M & 0.5B \\
    \midrule
    Muon & 346.57 & 699.15 & 1653.88 & 621.27 & 1560.48 & 2404.17 \\
    NorMuon & 346.73 & 699.51 & 1654.85 & 621.86 & 1561.53 & 2405.33 \\
    MeqMuon & \textbf{221.53} & \textbf{511.57} & \textbf{1403.69} & \textbf{513.13} & \textbf{1380.24} & \textbf{1884.59} \\
    \bottomrule
  \end{tabular}
\end{table}
\paragraph{Normalization direction.}

For the 2D weights in hidden layers, Table~\ref{tab:axis-choice} shows
MeqMuon's normalization-direction selection frequencies by module type.
The frequencies are aggregated over all training steps and layers for
Llama-350M and SmolLM2-360M.
The selections exhibit a strong relationship with matrix geometry: the tall
\texttt{gate\_proj} and \texttt{up\_proj} matrices always select row-wise
normalization, whereas the wide \texttt{down\_proj} matrices always select
column-wise normalization, and the wide \texttt{k\_proj} and
\texttt{v\_proj} matrices in SmolLM2-360M also predominantly select the
column-wise normalization. These selections agree with the one-sided imbalance
induced by orthogonalizing rectangular matrices, as discussed in
Section~\ref{sec:diagnostics}. For square matrices, the preference depends on
the module and model: \texttt{q\_proj} predominantly selects row-wise
normalization, while \texttt{o\_proj} predominantly selects column-wise
normalization in both models.

\begin{table}[!t]
  \caption{Row-wise and column-wise normalization selections during pretraining.}
  \label{tab:axis-choice}
  \centering
  \normalsize
  \setlength{\tabcolsep}{4.5pt}
  \renewcommand{\arraystretch}{1.12}
  \begin{tabular*}{\linewidth}{@{\extracolsep{\fill}}lcccccc@{}}
    \toprule
    & \multicolumn{3}{c}{Llama-350M} & \multicolumn{3}{c}{SmolLM2-360M} \\
    \cmidrule(lr){2-4}\cmidrule(lr){5-7}
    Module & Shape & Row (\%) & Col.\ (\%) & Shape & Row (\%) & Col.\ (\%) \\
    \midrule
    \texttt{q\_proj}    & $1024\times1024$ & \textbf{89.87} & 10.13
                        & $960\times960$ & \textbf{94.02} & 5.98 \\
    \texttt{k\_proj}    & $1024\times1024$ & \textbf{99.21} & 0.79
                        & $320\times960$ & 3.97 & \textbf{96.03} \\
    \texttt{v\_proj}    & $1024\times1024$ & \textbf{99.72} & 0.28
                        & $320\times960$ & 1.80 & \textbf{98.20} \\
    \texttt{o\_proj}    & $1024\times1024$ & 10.06 & \textbf{89.94}
                        & $960\times960$ & 18.85 & \textbf{81.15} \\
    \texttt{gate\_proj} & $2736\times1024$ & \textbf{100.00} & 0.00
                        & $2560\times960$ & \textbf{100.00} & 0.00 \\
    \texttt{up\_proj}   & $2736\times1024$ & \textbf{100.00} & 0.00
                        & $2560\times960$ & \textbf{100.00} & 0.00 \\
    \texttt{down\_proj} & $1024\times2736$ & 0.00 & \textbf{100.00}
                        & $960\times2560$ & 0.00 & \textbf{100.00} \\
    \bottomrule
  \end{tabular*}
\end{table}

\paragraph{Ablation study.}

We conduct component-wise ablations to
separate the contributions of MeqMuon across three parameter
types defined in Section~\ref{sec:hidden}. All variants start from Muon.
For 2D weights in hidden layers, the corresponding variant applies adaptive
one-sided normalization to their orthogonalized updates. For the remaining
2D parameters, the corresponding variant replaces AdamW with two-sided
normalization of their unorthogonalized momentum. For 1D parameters, the
corresponding variant replaces AdamW by normalizing each momentum vector by
its global RMS. Each single-component variant modifies only one parameter
group, while MeqMuon modifies all three.
Table~\ref{tab:experiment-ablation} shows the validation PPL and optimizer-state
memory. The variants that modify only 2D weights in hidden layers or only the
remaining 2D parameters both achieve lower validation PPL than Muon at both
model scales, while MeqMuon achieves the lowest~(best) PPL.
The optimizer-state memory savings come from replacing AdamW updates with normalized momentum for the remaining 2D
parameters and 1D parameters.

\begin{table}[!t]
  \caption{Component-wise ablation of validation PPL and optimizer-state
  memory on Llama models.}
  \label{tab:experiment-ablation}
  \centering
  \normalsize
  \renewcommand{\arraystretch}{1.12}
  \begin{tabular*}{\linewidth}{@{\extracolsep{\fill}}lcccc@{}}
    \toprule
    & \multicolumn{2}{c}{Llama-60M}
    & \multicolumn{2}{c}{Llama-130M} \\
    \cmidrule(lr){2-3}\cmidrule(lr){4-5}
    Variant & PPL ($\downarrow$) & Memory (MiB, $\downarrow$)
    & PPL ($\downarrow$) & Memory (MiB, $\downarrow$) \\
    \midrule
    Muon & 29.88 & 346.57 & 21.64 & 699.15 \\
    Only 2D weights in hidden layers & 29.74 & 346.57 & 21.53 & 699.15 \\
    Only remaining 2D parameters & 29.81 & 221.57 & 21.52 & 511.65 \\
    Only 1D parameters & 29.81 & 346.53 & 21.69 & 699.07 \\
    MeqMuon & \textbf{29.53} & \textbf{221.53}
    & \textbf{21.38} & \textbf{511.57} \\
    \bottomrule
  \end{tabular*}
\end{table}
\section{Conclusion}
\label{sec:conclusion}

In this paper, we propose an improved Muon optimizer, called \underline{m}atrix-\underline{eq}uilibrating Muon~(MeqMuon), for LLM pretraining.
MeqMuon balances both row and column magnitudes through normalization that can be automatically tailored to different imbalance patterns without manual intervention.
It eliminates second-moment storage by replacing AdamW updates with normalized momentum for the remaining
parameters.
Empirical results demonstrate that MeqMuon achieves better convergence
performance than AdamW, Muon, and other baselines in LLM pretraining.

\bibliography{references}

@article{chen2021evaluating,
  title={Evaluating large language models trained on code},
  author={Chen, Mark and Tworek, Jerry and Jun, Heewoo and Yuan, Qiming and Pinto, Henrique Ponde De Oliveira and Kaplan, Jared and Edwards, Harri and Burda, Yuri and Joseph, Nicholas and Brockman, Greg and others},
  journal={arXiv preprint arXiv:2107.03374},
  year={2021}
}

@article{deepseek2026v4,
  title={Deepseek-v4: Towards highly efficient million-token context intelligence},
  author={DeepSeek-AI},
  journal={arXiv preprint arXiv:2606.19348},
  year={2026}
}

@article{kimiteam2026k3,
  title={Kimi k3: Open frontier intelligence},
      author={{Kimi Team}},
  journal={arXiv preprint arXiv:2607.24653},
  year={2026}
}

@article{kimiteam2025k2,
      title={Kimi k2: Open agentic intelligence},
      author={{Kimi Team}},
      journal={arXiv preprint arXiv:2507.20534},
      year={2025}
}

@article{shao2025deepseekmathv2,
  title={Deepseekmath-v2: Towards self-verifiable mathematical reasoning},
  author={Shao, Zhihong and Luo, Yuxiang and Lu, Chengda and Ren, ZZ and Hu, Jiewen and Ye, Tian and Gou, Zhibin and Ma, Shirong and Zhang, Xiaokang},
  journal={arXiv preprint arXiv:2511.22570},
  year={2025}
}

@article{candido2026language,
  title={Language modeling materializes a world model of protein biology},
  author={Candido, Salvatore and Hayes, Thomas and Derry, Alexander and Rao, Roshan and Lin, Zeming and Verkuil, Robert and Wu, Bryan Z and Lee, Jin Sub and Bruguera, Elise S and Keval, Jehan A and others},
  journal={bioRxiv},
  year={2026},
}

@article{lin2023evolutionary,
  author       = {Zeming Lin and Halil Akin and Roshan Rao and Brian Hie and
                  Zhongkai Zhu and Wenting Lu and Nikita Smetanin and
                  Robert Verkuil and Ori Kabeli and Yaniv Shmueli and
                  Allan {dos Santos Costa} and Maryam Fazel-Zarandi and
                  Tom Sercu and Salvatore Candido and Alexander Rives},
  title        = {Evolutionary-scale prediction of atomic-level protein
                  structure with a language model},
  journal      = {Science},
  volume       = {379},
  number       = {6637},
  pages        = {1123--1130},
  year         = {2023},
}

@misc{jordan2024muon,
  author       = {Keller Jordan and Yuchen Jin and Vlado Boza and Jiacheng You and
                  Franz Cesista and Laker Newhouse and Jeremy Bernstein},
  title        = {Muon: An optimizer for hidden layers in neural networks},
  year         = {2024},
  url          = {https://kellerjordan.github.io/posts/muon/}
}

@article{liu2025scalable,
  title={Muon is scalable for llm training},
  author={Liu, Jingyuan and Su, Jianlin and Yao, Xingcheng and Jiang, Zhejun and Lai, Guokun and Du, Yulun and Qin, Yidao and Xu, Weixin and Lu, Enzhe and Yan, Junjie and others},
  journal={arXiv preprint arXiv:2502.16982},
  year={2025}
}

@inproceedings{li2025normuon,
  author       = {Zichong Li and Liming Liu and Chen Liang and Weizhu Chen and Tuo Zhao},
  title        = {{NorMuon}: Making {Muon} More Efficient and Scalable},
  booktitle    = {Proceedings of the International Conference on Machine Learning},
  year         = {2026},
}

@inproceedings{glentis2026scale,
  author       = {Athanasios Glentis and Jiaxiang Li and Andi Han and Mingyi Hong},
  title        = {Memory-Efficient {LLM} Pretraining via Minimalist Optimizer Design},
  booktitle    = {Proceedings of the International Conference on Machine Learning},
  year         = {2026},
}

@article{levy2016power,
  title={The power of normalization: Faster evasion of saddle points},
  author={Levy, Kfir Y},
  journal={arXiv preprint arXiv:1611.04831},
  year={2016}
}

@article{murray2019revisiting,
  author={Murray, Ryan and Swenson, Brian and Kar, Soummya},
  journal={IEEE Transactions on Automatic Control}, 
  title={Revisiting Normalized Gradient Descent: Fast Evasion of Saddle Points}, 
  year={2019},
  volume={64},
  number={11},
  pages={4818-4824},
}

@inproceedings{cutkosky2020momentum,
  title = 	 {Momentum Improves Normalized {SGD}},
  author =       {Cutkosky, Ashok and Mehta, Harsh},
  booktitle = 	 {Proceedings of the International Conference on Machine Learning},
  year = 	 {2020},
}

@article{zhao2024stochastic,
  author       = {Shen-Yi Zhao and Chang-Wei Shi and Yin-Peng Xie and Wu-Jun Li},
  title        = {Stochastic Normalized Gradient Descent with Momentum for
                  Large-Batch Training},
  journal      = {Science China Information Sciences},
  volume       = {67},
  number       ={11},
  pages        = {212101},
  year         = {2024},
}

@inproceedings{yang2024batch,
 author = {Yang, Yi-Rui and Shi, Chang-Wei and Li, Wu-Jun},
 booktitle = {Proceedings of the International Conference on Learning Representations},
 title = {On the Effect of Batch Size in Byzantine-Robust Distributed Learning},
 year = {2024}
}

@inproceedings{zhang2020clipping,
  author       = {Jingzhao Zhang and Tianxing He and Suvrit Sra and Ali Jadbabaie},
  title        = {Why Gradient Clipping Accelerates Training:
                  A Theoretical Justification for Adaptivity},
  booktitle    = {Proceedings of the International Conference on Learning Representations},
  year         = {2020},
}

@article{sun2025revisiting,
  author  = {Tao Sun and Xinwang Liu and Kun Yuan},
  title   = {Revisiting Gradient Normalization and Clipping for Nonconvex SGD under Heavy-Tailed Noise: Necessity, Sufficiency, and Acceleration},
  journal = {Journal of Machine Learning Research},
  year    = {2025},
  volume  = {26},
  number  = {237},
  pages   = {1--42},
}

@inproceedings{kingma2015adam,
  author       = {Diederik P. Kingma and Jimmy Ba},
  title        = {{Adam}: A Method for Stochastic Optimization},
  booktitle    = {Proceedings of the International Conference on Learning Representations},
  year         = {2015},
}

@inproceedings{pagliardini2025ademamix,
 author = {Pagliardini, Matteo and Ablin, Pierre and Grangier, David},
 booktitle = {Proceedings of the International Conference on Learning Representations},
 title = {The AdEMAMix Optimizer: Better, Faster, Older},
 year = {2025}
}

@inproceedings{dettmers2022eightbit,
  author       = {Tim Dettmers and Mike Lewis and Sam Shleifer and Luke Zettlemoyer},
  title        = {8-bit Optimizers via Block-wise Quantization},
  booktitle    = {Proceedings of the International Conference on Learning Representations},
  year         = {2022},
}

@inproceedings{shazeer2018adafactor,
  title = 	 {Adafactor: Adaptive Learning Rates with Sublinear Memory Cost},
  author =       {Shazeer, Noam and Stern, Mitchell},
  booktitle = 	 {Proceedings of the International Conference on Machine Learning},
  year = 	 {2018},
}

@inproceedings{zhang2025adammini,
 author = {Zhang, Yushun and Chen, Congliang and Li, Ziniu and Ding, Tian and Wu, Chenwei and Kingma, Diederik (Durk) and Ye, Yinyu and Luo, Zhi-Quan and Sun, Ruoyu},
 booktitle = {Proceedings of the International Conference on Learning Representations},
 title = {Adam-mini: Use Fewer Learning Rates To Gain More},
 year = {2025}
}

@inproceedings{zhao2024galore,
  title = 	 {{G}a{L}ore: Memory-Efficient {LLM} Training by Gradient Low-Rank Projection},
  author =       {Zhao, Jiawei and Zhang, Zhenyu and Chen, Beidi and Wang, Zhangyang and Anandkumar, Anima and Tian, Yuandong},
  booktitle = 	 {Proceedings of the International Conference on Machine Learning},
  year = 	 {2024},
}

@inproceedings{zhu2025apollo,
 author = {Zhu, Hanqing and Zhang, Zhenyu and Cong, Wenyan and Liu, Xi and Park, Sem and Chandra, Vikas and Long, Bo and Pan, David Z. and Wang, Zhangyang and Lee, Jinwon},
 booktitle = {Proceedings of Machine Learning and Systems},
 title = {APOLLO: SGD-like Memory, AdamW-level Performance},
 year = {2025}
}

@inproceedings{chen2023lion,
 author = {Chen, Xiangning and Liang, Chen and Huang, Da and Real, Esteban and Wang, Kaiyuan and Pham, Hieu and Dong, Xuanyi and Luong, Thang and Hsieh, Cho-Jui and Lu, Yifeng and Le, Quoc V},
 booktitle = {Advances in Neural Information Processing Systems},
 pages = {49205--49233},
 title = {Symbolic Discovery of Optimization Algorithms},
 year = {2023}
}

@inproceedings{loshchilov2019adamw,
  author       = {Ilya Loshchilov and Frank Hutter},
  title        = {Decoupled Weight Decay Regularization},
  booktitle    = {Proceedings of the International Conference on Learning Representations},
  year         = {2019},
}

@article{raffel2020t5,
  author  = {Colin Raffel and Noam Shazeer and Adam Roberts and Katherine Lee and Sharan Narang and Michael Matena and Yanqi Zhou and Wei Li and Peter J. Liu},
  title   = {Exploring the Limits of Transfer Learning with a Unified Text-to-Text Transformer},
  journal = {Journal of Machine Learning Research},
  year    = {2020},
  volume  = {21},
  number  = {140},
  pages   = {1-67},
}

@article{benallal2025smollm2,
  title={SmolLM2: When Smol Goes Big--Data-Centric Training of a Small Language Model},
  author={Allal, Loubna Ben and Lozhkov, Anton and Bakouch, Elie and Bl{\'a}zquez, Gabriel Mart{\'\i}n and Penedo, Guilherme and Tunstall, Lewis and Marafioti, Andr{\'e}s and Kydl{\'\i}{\v{c}}ek, Hynek and Lajar{\'\i}n, Agust{\'\i}n Piqueres and Srivastav, Vaibhav and others},
  journal={arXiv preprint arXiv:2502.02737},
  year={2025}
}

@article{deepseek2024v3,
  title={Deepseek-v3 technical report},
  author={Liu, Aixin and Feng, Bei and Xue, Bing and Wang, Bingxuan and Wu, Bochao and Lu, Chengda and Zhao, Chenggang and Deng, Chengqi and Zhang, Chenyu and Ruan, Chong and others},
  journal={arXiv preprint arXiv:2412.19437},
  year={2024}
}

@article{touvron2023llama2,
  title={Llama 2: Open foundation and fine-tuned chat models},
  author={Touvron, Hugo and Martin, Louis and Stone, Kevin and Albert, Peter and Almahairi, Amjad and Babaei, Yasmine and Bashlykov, Nikolay and Batra, Soumya and Bhargava, Prajjwal and Bhosale, Shruti and others},
  journal={arXiv preprint arXiv:2307.09288},
  year={2023}
}

@article{hoffmann2022training,
  title={Training compute-optimal large language models},
  author={Hoffmann, Jordan and Borgeaud, Sebastian and Mensch, Arthur and Buchatskaya, Elena and Cai, Trevor and Rutherford, Eliza and Casas, Diego de Las and Hendricks, Lisa Anne and Welbl, Johannes and Clark, Aidan and others},
  journal={arXiv preprint arXiv:2203.15556},
  year={2022}
}

@article{yang2024qwen2,
      title={Qwen2 Technical Report}, 
      author={An Yang and Baosong Yang and Binyuan Hui and Bo Zheng and Bowen Yu and Chang Zhou and Chengpeng Li and Chengyuan Li and Dayiheng Liu and Fei Huang and Guanting Dong and Haoran Wei and Huan Lin and Jialong Tang and Jialin Wang and Jian Yang and Jianhong Tu and Jianwei Zhang and Jianxin Ma and Jianxin Yang and Jin Xu and Jingren Zhou and Jinze Bai and Jinzheng He and Junyang Lin and Kai Dang and Keming Lu and Keqin Chen and Kexin Yang and Mei Li and Mingfeng Xue and Na Ni and Pei Zhang and Peng Wang and Ru Peng and Rui Men and Ruize Gao and Runji Lin and Shijie Wang and Shuai Bai and Sinan Tan and Tianhang Zhu and Tianhao Li and Tianyu Liu and Wenbin Ge and Xiaodong Deng and Xiaohuan Zhou and Xingzhang Ren and Xinyu Zhang and Xipin Wei and Xuancheng Ren and Xuejing Liu and Yang Fan and Yang Yao and Yichang Zhang and Yu Wan and Yunfei Chu and Yuqiong Liu and Zeyu Cui and Zhenru Zhang and Zhifang Guo and Zhihao Fan},
      journal={arXiv preprint arXiv:2407.10671},
      year={2024},      
}

@inproceedings{zhang2019rmsnorm,
 author = {Zhang, Biao and Sennrich, Rico},
 booktitle = {Advances in Neural Information Processing Systems},
 title = {Root Mean Square Layer Normalization},
 year = {2019}
}
\bibliographystyle{iclr2027_conference}

\newpage
\appendix
\section{Additional imbalance patterns}
\label{app:imbalance}

Table~\ref{tab:appendix-cv-llama350}, Table~\ref{tab:appendix-cv-llama60},
and Table~\ref{tab:appendix-cv-smollm135} report the row and column CVs,
$\gamma_r$ and $\gamma_c$ defined in \eqref{eq:cv}, of
Muon's orthogonalized updates $\mU_t$ for Llama-350M, Llama-60M, and
SmolLM2-135M, respectively.
For each model, we select three different training steps and
three Transformer layers.
The tables report all seven update matrices for 2D weights in hidden layers after 5 NS iterations.
For each training step, the larger of $\gamma_r$ and $\gamma_c$ is shown in bold.

\begin{table}[!t]
  \caption{Row and column CVs for Llama-350M after 5 NS iterations.}
  \label{tab:appendix-cv-llama350}
  \centering
  \small
  \setlength{\tabcolsep}{4.5pt}
  \renewcommand{\arraystretch}{1.25}
  \setlength{\aboverulesep}{0pt}
  \setlength{\belowrulesep}{0pt}
  \begin{tabular}{clccc|cc|cc}
    \toprule
    & & & \multicolumn{2}{c|}{Step 381} & \multicolumn{2}{c|}{Step 1144} & \multicolumn{2}{c}{Step 1907} \\
    \cmidrule(lr){4-5}\cmidrule(lr){6-7}\cmidrule(lr){8-9}
    Layer & Matrix & Shape & $\gamma_r$ & $\gamma_c$ & $\gamma_r$ & $\gamma_c$ & $\gamma_r$ & $\gamma_c$ \\
    \midrule
    \multirow{7}{*}{0} & \texttt{q\_proj} & $1024 \times 1024$ & \textbf{0.033} & 0.012 & \textbf{0.084} & 0.022 & \textbf{0.066} & 0.024 \\
     & \texttt{k\_proj} & $1024 \times 1024$ & \textbf{0.026} & 0.012 & \textbf{0.079} & 0.022 & \textbf{0.073} & 0.027 \\
     & \texttt{v\_proj} & $1024 \times 1024$ & \textbf{0.078} & 0.019 & \textbf{0.084} & 0.027 & \textbf{0.082} & 0.024 \\
     & \texttt{o\_proj} & $1024 \times 1024$ & 0.015 & \textbf{0.063} & 0.013 & \textbf{0.053} & 0.013 & \textbf{0.048} \\
     & \texttt{gate\_proj} & $2736 \times 1024$ & \textbf{0.603} & 0.020 & \textbf{0.534} & 0.026 & \textbf{0.528} & 0.023 \\
     & \texttt{up\_proj} & $2736 \times 1024$ & \textbf{0.616} & 0.020 & \textbf{0.538} & 0.026 & \textbf{0.531} & 0.023 \\
     & \texttt{down\_proj} & $1024 \times 2736$ & 0.009 & \textbf{0.558} & 0.013 & \textbf{0.505} & 0.014 & \textbf{0.505} \\
    \midrule
    \multirow{7}{*}{12} & \texttt{q\_proj} & $1024 \times 1024$ & \textbf{0.492} & 0.023 & \textbf{0.333} & 0.021 & \textbf{0.287} & 0.018 \\
     & \texttt{k\_proj} & $1024 \times 1024$ & \textbf{0.536} & 0.025 & \textbf{0.365} & 0.022 & \textbf{0.326} & 0.020 \\
     & \texttt{v\_proj} & $1024 \times 1024$ & \textbf{0.368} & 0.023 & \textbf{0.238} & 0.021 & \textbf{0.163} & 0.017 \\
     & \texttt{o\_proj} & $1024 \times 1024$ & 0.017 & \textbf{0.231} & 0.014 & \textbf{0.151} & 0.014 & \textbf{0.115} \\
     & \texttt{gate\_proj} & $2736 \times 1024$ & \textbf{0.248} & 0.013 & \textbf{0.155} & 0.015 & \textbf{0.132} & 0.016 \\
     & \texttt{up\_proj} & $2736 \times 1024$ & \textbf{0.227} & 0.014 & \textbf{0.147} & 0.014 & \textbf{0.129} & 0.016 \\
     & \texttt{down\_proj} & $1024 \times 2736$ & 0.014 & \textbf{0.250} & 0.014 & \textbf{0.150} & 0.015 & \textbf{0.119} \\
    \midrule
    \multirow{7}{*}{23} & \texttt{q\_proj} & $1024 \times 1024$ & \textbf{0.218} & 0.014 & \textbf{0.050} & 0.011 & \textbf{0.038} & 0.011 \\
     & \texttt{k\_proj} & $1024 \times 1024$ & \textbf{0.234} & 0.016 & \textbf{0.056} & 0.011 & \textbf{0.043} & 0.011 \\
     & \texttt{v\_proj} & $1024 \times 1024$ & \textbf{0.207} & 0.019 & \textbf{0.055} & 0.012 & \textbf{0.043} & 0.011 \\
     & \texttt{o\_proj} & $1024 \times 1024$ & 0.025 & \textbf{0.187} & 0.014 & \textbf{0.060} & 0.013 & \textbf{0.043} \\
     & \texttt{gate\_proj} & $2736 \times 1024$ & \textbf{0.315} & 0.013 & \textbf{0.214} & 0.015 & \textbf{0.198} & 0.016 \\
     & \texttt{up\_proj} & $2736 \times 1024$ & \textbf{0.317} & 0.013 & \textbf{0.230} & 0.014 & \textbf{0.214} & 0.015 \\
     & \texttt{down\_proj} & $1024 \times 2736$ & 0.020 & \textbf{0.303} & 0.018 & \textbf{0.205} & 0.017 & \textbf{0.187} \\
    \bottomrule
  \end{tabular}
\end{table}

\begin{table}[!t]
  \caption{Row and column CVs for Llama-60M after 5 NS iterations.}
  \label{tab:appendix-cv-llama60}
  \centering
  \small
  \setlength{\tabcolsep}{4.5pt}
  \renewcommand{\arraystretch}{1.25}
  \setlength{\aboverulesep}{0pt}
  \setlength{\belowrulesep}{0pt}
  \begin{tabular}{clccc|cc|cc}
    \toprule
    & & & \multicolumn{2}{c|}{Step 443} & \multicolumn{2}{c|}{Step 1329} & \multicolumn{2}{c}{Step 2215} \\
    \cmidrule(lr){4-5}\cmidrule(lr){6-7}\cmidrule(lr){8-9}
    Layer & Matrix & Shape & $\gamma_r$ & $\gamma_c$ & $\gamma_r$ & $\gamma_c$ & $\gamma_r$ & $\gamma_c$ \\
    \midrule
    \multirow{7}{*}{0} & \texttt{q\_proj} & $512 \times 512$ & \textbf{0.034} & 0.019 & \textbf{0.043} & 0.023 & \textbf{0.040} & 0.021 \\
     & \texttt{k\_proj} & $512 \times 512$ & \textbf{0.088} & 0.031 & \textbf{0.076} & 0.031 & \textbf{0.066} & 0.026 \\
     & \texttt{v\_proj} & $512 \times 512$ & \textbf{0.164} & 0.081 & \textbf{0.101} & 0.041 & \textbf{0.066} & 0.025 \\
     & \texttt{o\_proj} & $512 \times 512$ & 0.034 & \textbf{0.144} & 0.020 & \textbf{0.087} & 0.017 & \textbf{0.050} \\
     & \texttt{gate\_proj} & $1376 \times 512$ & \textbf{0.654} & 0.022 & \textbf{0.528} & 0.017 & \textbf{0.503} & 0.017 \\
     & \texttt{up\_proj} & $1376 \times 512$ & \textbf{0.665} & 0.026 & \textbf{0.522} & 0.013 & \textbf{0.497} & 0.016 \\
     & \texttt{down\_proj} & $512 \times 1376$ & 0.012 & \textbf{0.603} & 0.016 & \textbf{0.523} & 0.016 & \textbf{0.506} \\
    \midrule
    \multirow{7}{*}{4} & \texttt{q\_proj} & $512 \times 512$ & \textbf{0.167} & 0.026 & \textbf{0.027} & 0.015 & \textbf{0.023} & 0.014 \\
     & \texttt{k\_proj} & $512 \times 512$ & \textbf{0.206} & 0.026 & \textbf{0.029} & 0.015 & \textbf{0.027} & 0.015 \\
     & \texttt{v\_proj} & $512 \times 512$ & \textbf{0.145} & 0.029 & \textbf{0.032} & 0.018 & \textbf{0.019} & 0.016 \\
     & \texttt{o\_proj} & $512 \times 512$ & 0.017 & \textbf{0.041} & 0.014 & \textbf{0.017} & 0.014 & \textbf{0.016} \\
     & \texttt{gate\_proj} & $1376 \times 512$ & \textbf{0.128} & 0.016 & \textbf{0.075} & 0.017 & \textbf{0.067} & 0.017 \\
     & \texttt{up\_proj} & $1376 \times 512$ & \textbf{0.127} & 0.016 & \textbf{0.086} & 0.018 & \textbf{0.082} & 0.016 \\
     & \texttt{down\_proj} & $512 \times 1376$ & 0.017 & \textbf{0.128} & 0.016 & \textbf{0.084} & 0.016 & \textbf{0.077} \\
    \midrule
    \multirow{7}{*}{7} & \texttt{q\_proj} & $512 \times 512$ & \textbf{0.059} & 0.015 & \textbf{0.026} & 0.015 & \textbf{0.026} & 0.014 \\
     & \texttt{k\_proj} & $512 \times 512$ & \textbf{0.073} & 0.017 & \textbf{0.030} & 0.014 & \textbf{0.029} & 0.014 \\
     & \texttt{v\_proj} & $512 \times 512$ & \textbf{0.052} & 0.021 & \textbf{0.018} & 0.015 & \textbf{0.019} & 0.015 \\
     & \texttt{o\_proj} & $512 \times 512$ & 0.019 & \textbf{0.038} & 0.016 & \textbf{0.019} & 0.016 & \textbf{0.016} \\
     & \texttt{gate\_proj} & $1376 \times 512$ & \textbf{0.144} & 0.018 & \textbf{0.093} & 0.019 & \textbf{0.092} & 0.016 \\
     & \texttt{up\_proj} & $1376 \times 512$ & \textbf{0.167} & 0.016 & \textbf{0.123} & 0.018 & \textbf{0.120} & 0.017 \\
     & \texttt{down\_proj} & $512 \times 1376$ & 0.019 & \textbf{0.136} & 0.020 & \textbf{0.093} & 0.021 & \textbf{0.098} \\
    \bottomrule
  \end{tabular}
\end{table}

\begin{table}[!t]
  \caption{Row and column CVs for SmolLM2-135M after 5 NS iterations.}
  \label{tab:appendix-cv-smollm135}
  \centering
  \small
  \setlength{\tabcolsep}{4.5pt}
  \renewcommand{\arraystretch}{1.25}
  \setlength{\aboverulesep}{0pt}
  \setlength{\belowrulesep}{0pt}
  \begin{tabular}{clccc|cc|cc}
    \toprule
    & & & \multicolumn{2}{c|}{Step 381} & \multicolumn{2}{c|}{Step 1144} & \multicolumn{2}{c}{Step 1907} \\
    \cmidrule(lr){4-5}\cmidrule(lr){6-7}\cmidrule(lr){8-9}
    Layer & Matrix & Shape & $\gamma_r$ & $\gamma_c$ & $\gamma_r$ & $\gamma_c$ & $\gamma_r$ & $\gamma_c$ \\
    \midrule
    \multirow{7}{*}{0} & \texttt{q\_proj} & $576 \times 576$ & \textbf{0.018} & 0.014 & \textbf{0.033} & 0.015 & \textbf{0.026} & 0.013 \\
     & \texttt{k\_proj} & $192 \times 576$ & 0.023 & \textbf{0.084} & 0.027 & \textbf{0.089} & 0.030 & \textbf{0.090} \\
     & \texttt{v\_proj} & $192 \times 576$ & 0.020 & \textbf{0.082} & 0.020 & \textbf{0.084} & 0.020 & \textbf{0.082} \\
     & \texttt{o\_proj} & $576 \times 576$ & 0.019 & \textbf{0.038} & 0.016 & \textbf{0.027} & 0.015 & \textbf{0.026} \\
     & \texttt{gate\_proj} & $1536 \times 576$ & \textbf{0.421} & 0.023 & \textbf{0.370} & 0.021 & \textbf{0.363} & 0.019 \\
     & \texttt{up\_proj} & $1536 \times 576$ & \textbf{0.434} & 0.024 & \textbf{0.370} & 0.021 & \textbf{0.363} & 0.019 \\
     & \texttt{down\_proj} & $576 \times 1536$ & 0.012 & \textbf{0.410} & 0.015 & \textbf{0.353} & 0.015 & \textbf{0.348} \\
    \midrule
    \multirow{7}{*}{15} & \texttt{q\_proj} & $576 \times 576$ & \textbf{0.033} & 0.013 & \textbf{0.040} & 0.013 & \textbf{0.029} & 0.013 \\
     & \texttt{k\_proj} & $192 \times 576$ & 0.038 & \textbf{0.063} & 0.031 & \textbf{0.061} & 0.049 & \textbf{0.063} \\
     & \texttt{v\_proj} & $192 \times 576$ & 0.025 & \textbf{0.062} & 0.042 & \textbf{0.068} & 0.039 & \textbf{0.064} \\
     & \texttt{o\_proj} & $576 \times 576$ & 0.014 & \textbf{0.017} & 0.014 & \textbf{0.017} & 0.014 & \textbf{0.014} \\
     & \texttt{gate\_proj} & $1536 \times 576$ & \textbf{0.248} & 0.017 & \textbf{0.122} & 0.018 & \textbf{0.096} & 0.017 \\
     & \texttt{up\_proj} & $1536 \times 576$ & \textbf{0.232} & 0.016 & \textbf{0.125} & 0.018 & \textbf{0.105} & 0.018 \\
     & \texttt{down\_proj} & $576 \times 1536$ & 0.016 & \textbf{0.231} & 0.016 & \textbf{0.116} & 0.016 & \textbf{0.090} \\
    \midrule
    \multirow{7}{*}{29} & \texttt{q\_proj} & $576 \times 576$ & \textbf{0.017} & 0.015 & \textbf{0.021} & 0.015 & \textbf{0.020} & 0.013 \\
     & \texttt{k\_proj} & $192 \times 576$ & 0.028 & \textbf{0.076} & 0.029 & \textbf{0.077} & 0.027 & \textbf{0.071} \\
     & \texttt{v\_proj} & $192 \times 576$ & 0.026 & \textbf{0.083} & 0.025 & \textbf{0.078} & 0.023 & \textbf{0.067} \\
     & \texttt{o\_proj} & $576 \times 576$ & 0.018 & \textbf{0.019} & 0.015 & \textbf{0.015} & \textbf{0.017} & 0.014 \\
     & \texttt{gate\_proj} & $1536 \times 576$ & \textbf{0.321} & 0.017 & \textbf{0.206} & 0.017 & \textbf{0.190} & 0.017 \\
     & \texttt{up\_proj} & $1536 \times 576$ & \textbf{0.319} & 0.017 & \textbf{0.210} & 0.019 & \textbf{0.193} & 0.018 \\
     & \texttt{down\_proj} & $576 \times 1536$ & 0.019 & \textbf{0.302} & 0.022 & \textbf{0.198} & 0.023 & \textbf{0.181} \\
    \bottomrule
  \end{tabular}
\end{table}

\section{Model Architectures}
\label{app:implementation-settings}

Table~\ref{tab:experiment-architecture} gives the architecture of each
model. All models use pre-normalization with RMSNorm, SwiGLU feed-forward
layers, and rotary position embeddings. The Llama models use multi-head
attention, whereas SmolLM2 and Qwen2 use grouped-query attention.
Qwen2 additionally uses biases in the query, key, and value projections.
The Params column gives the number of unique parameters in millions, so tied token-embedding and
LM-head weights are counted once. Q/KV gives the numbers of query and
key/value heads, and Tied indicates whether the token embedding and the
LM head share weights.

\begin{table}[!t]
  \caption{Model architectures.}
  \label{tab:experiment-architecture}
  \centering
  \normalsize
  \setlength{\tabcolsep}{4.5pt}
  \renewcommand{\arraystretch}{1.12}
  \begin{tabular}{llrrrrcrrc}
    \toprule
    Family & Size & Params (M) & Layers & Hidden & FFN & Q/KV
      & Head dim & Vocab. & Tied \\
    \midrule
    \multirow{3}{*}{Llama}
      & 60M  & 58.074  & 8  & 512  & 1376 & 8/8   & 64 & 32,000 & No \\
      & 130M & 134.106 & 12 & 768  & 2048 & 12/12 & 64 & 32,000 & No \\
      & 350M & 367.969 & 24 & 1024 & 2736 & 16/16 & 64 & 32,000 & No \\
    \midrule
    \multirow{2}{*}{SmolLM2}
      & 135M & 134.515 & 30 & 576  & 1536 & 9/3   & 64 & 49,152 & Yes \\
      & 360M & 361.821 & 32 & 960  & 2560 & 15/5  & 64 & 49,152 & Yes \\
    \midrule
    Qwen2
      & 0.5B & 494.033 & 24 & 896  & 4864 & 14/2  & 64 & 151,936 & Yes \\
    \bottomrule
  \end{tabular}
\end{table}

\end{document}